\documentclass[11pt,twoside]{article}
\usepackage{a4wide,charter}
\usepackage[T1]{fontenc}
\usepackage{amssymb}
\usepackage{framed}
\usepackage[normalem]{ulem}

\usepackage{fancyhdr}
\renewcommand{\headheight}{14.5pt}

\usepackage[english]{babel}
\usepackage[utf8]{inputenc}
\usepackage{amsmath}
\usepackage{graphicx}
\usepackage[colorinlistoftodos]{todonotes}

\usepackage[numbers]{natbib}
\usepackage{hyperref}

\title{Evolutionary Innovations and Where to Find Them\\
  \vspace{1mm}\Large{\emph{Routes to Open-Ended Evolution in Natural
  and Artificial Systems}}}
\author{Tim Taylor$^{1,2,3}$\\ 
\mbox{}\\
$^1$Independent Researcher, Edinburgh, U.K.\\
$^2$Faculty of Information Technology, Monash University, Melbourne, Australia\\
$^3$International Programmes in Computing, University of London, U.K.\\
tim@tim-taylor.com}

\date{}

\begin{document}

\maketitle

\begin{abstract}
This paper presents a high-level conceptual framework to help orient the
discussion and implementation of open-endedness in evolutionary
systems. Drawing upon earlier work by Banzhaf \emph{et al}., three
different kinds of open-endedness are identified: \emph{exploratory},
\emph{expansive}, and \emph{transformational}. These are characterised
in terms of their relationship to the search space of phenotypic
behaviours. A formalism is introduced to describe three key processes
required for an evolutionary process: the generation of a phenotype
from a genetic description, the evaluation of that phenotype, and the
reproduction with variation of individuals according to their
evaluation. The formalism makes explicit various influences in each of
these processes that can easily be overlooked. The distinction is made
between intrinsic and extrinsic implementations of these processes. A
discussion then investigates how various interactions between these 
processes, and their modes of implementation, can lead to
open-endedness.
However, an important contribution of the paper is
the demonstration that these considerations
relate to exploratory open-endedness only. Conditions for the
implementation of the more interesting kinds of
open-endedness---expansive and transformational---are also discussed,
emphasizing factors such as multiple domains of behaviour, transdomain
bridges, and non-additive compositional systems. In
contrast to a traditional Darwinian analysis, these factors
relate not to the generic evolutionary properties of individuals and
populations, but rather to the nature of the building blocks out of
which individual organisms are constructed, and the laws and
properties of the environment in which they exist. The paper ends with
suggestions of how the framework can be used to categorise and compare
the open-ended evolutionary potential of different systems, how it
might guide the design of systems with greater capacity for open-ended
evolution, and how it might be further improved.
\end{abstract}

\section{Introduction}

In this paper I identify different routes by which open-endedness (OE) can
be introduced into the design and implementation of an evolutionary system.

I begin by presenting a definition of three different kinds of
open-endedness.  My treatment of the topic expands upon
the approach recently proposed by Banzhaf \emph{et al}.\
\cite{Banzhaf:Defining}. 
In their paper, the distinction is made between
\emph{scientific models}, which are  
``descriptive models of part of the existing world'', and
\emph{engineering models} (including software design models), which
are ``prescriptive or normative models of a system to be built in the
world'' \citep[p.\ 135]{Banzhaf:Defining}. One of the main aims of
their paper was to 
develop a descriptive scientific (meta-)model to illustrate their
definitions of open-endedness. They express the hope that ``such a
definition of OE in terms of models and meta-models will help the
design of normative engineering models for implementing ALife'' \citep[p.\
136]{Banzhaf:Defining}.

The aim of the current contribution is to make
progress towards exactly that goal---the development of an engineering
model to guide the design and implementation of artificial
evolutionary systems that possess the capacity for various kinds of
open-endedness.\footnote{This paper concentrates
  specifically on open-ended \emph{evolution}. In places reference is
  made to the more general concept of \emph{open-endedness} (which admits
  that open-ended dynamics may be observed in other types of system
  beyond evolutionary ones \cite{Stanley:OE}), but the discussion
  presented here assumes an evolutionary context.}

Having clarified what I mean by open-endedness, I then introduce a
formalism for describing the key processes that must be present in
any evolutionary system. The formalism makes explicit some important
dependencies and interrelationships that are otherwise easy to
overlook. 

Equipped with the necessary preliminaries, I then utilise the
formalism to identify the various routes by which
open-endedness can be accommodated in the design of an evolutionary
system. It is found that the formalism only helps directly in the
investigation of one type of open-endedness. I therefore continue the
discussion with an analysis of potential factors involved in the other
kinds of open-endedness as well.

Throughout the paper I demonstrate how the presented
framework\footnote{Note that throughout the paper I use the term
  ``framework'' to describe the whole approach to understanding
  open-endedness outlined here, which includes the definition of three
  kinds of open-ended evolution (OEE) (Section~\ref{sct-novelties}),
  the formalism describing the three basic evolutionary processes
  (Sections~\ref{sct-formalism} and 
  \ref{sct-explore-oe}), and the discussion of mechanisms for expansive 
  and transformational OEE (Section~\ref{sct-expans-trans-oe}). I use
  the term ``formalism'' to refer specifically to the subset of the
  framework relating to the basic evolutionary processes
  (Sections~\ref{sct-formalism} and \ref{sct-explore-oe}).}
helps orient the study of open-endedness within the context of existing
literature. I close the discussion by outlining how the framework
could be used as a tool for analysing and improving the open-endedness
of existing artificial evolutionary systems, and offering some
suggestions for further developments of the approach.

I consider the main contributions of the paper to be the analysis of
how the many different topics from the theoretical biology literature
fit into the overall picture of open-ended evolution (as summarised in
Figure~\ref{fig-routes}), the finding that these only relate to one
type of OE (Section~\ref{sct-explore-oe}), the discussion of ways of
achieving the other types of OE (Section~\ref{sct-expans-trans-oe}),
the suggestions for extending the analysis with a more sophisticated
treatment of behaviour (Section~\ref{sct-final-remarks}),
and also the simple schematic representation of Banzhaf \emph{et
  al}.’s classes of open-endedness, upon which the discussion is based
(Figure~\ref{fig-typesOfOE}).

\section{State Spaces, Novelties and Open-Endedness}
\label{sct-novelties}

The idea of a \emph{possibility space} or \emph{state space} to
represent the range of all possible forms of an individual in an
evolutionary system is a widely employed concept
 (e.g.\ \citep{deVladar:GrandViews, Boden:CreativityALife, Banzhaf:Defining}).
Indeed, they are simplifications of the the concept of \emph{adaptive
  landscapes} first proposed by Sewall Wright (for genotypes) in
1932 and by G.G.\ Simpson (for phenotypes) in 1944
\citep{Dietrich:Shifting}. 
State spaces are simpler than adaptive landscapes because they lack a
representation of the \emph{adaptive value (fitness)} of each point in
the space.
I use the simpler concept of state space in the following discussion
as it is sufficient for the purpose of the discussion; I consider how
fitness comes into the picture later in the paper.

While it is easy to use state spaces and adaptive landscapes to describe
particular, well constrained systems comprising a small number of
clearly defined variables, it is non-trivial to apply them to
elaborate and potentially open-ended systems. In these cases it can be
problematic to enumerate and quantify all relevant variables to be
used as dimensions of the space.\footnote{But note that methods for
  inferring latent variable models can be employed to
  generate more meaningful \emph{latent spaces}.} However, even if it
can be difficult to quantitatively describe a specific complex
evolutionary system, state spaces can still be useful \emph{intuition
  pumps} \citep{Dennett:IntuitionPumps}---this is my intention in
using them here.

To present the following ideas in more concrete terms, I have chosen
to illustrate state spaces defined according to the ideas of \emph{models}
and \emph{meta-models} set out in Banzhaf \emph{et al}.'s recent treatment
of open-endedness \cite{Banzhaf:Defining}. Central to their approach 
is the idea that the behaviour of a system can be described by a
scientific (descriptive) model. The model is expressed in terms of a
set of concepts, and those concepts can themselves be described by a
meta-model. The meta-model describes a set of concepts that can
be used to build a variety of specific models that use the same
concepts in different ways. Readers unfamiliar with Banzhaf \emph{et
  al}.'s contribution may benefit from reading it in order to fully
understand what is summarised in this section.

Banzhaf \emph{et al}.\ identify three different kinds of novelty that
may occur in a system, defined according to 
whether the novelty necessitates changes in the system's model or
meta-model. Their approach closely resembles Boden's ideas of
\emph{three different kinds of creativity} that have been developed
over several decades \citep{Boden:CreativeMind,Boden:CreativityALife}.
The distinction between novelties that fall within the system's
current model and those that necessitate a change in the model (or
meta-model) can also be seen in Waddington's pioneering work on
open-ended evolution from 1969:
\begin{quotation}
``the \ldots\ requirement, that the available genotypes must be
capable of producing phenotypes which can exploit \ldots\ new
environments, requires some special provision of a means of creating
genetic variation \ldots\ It is important to emphasize that the new
genetic variation must not only be novel, but must include variations
which make possible the exploration of environments which the
population previously did not utilize \ldots\ It is not sufficient to
produce new mutations which merely insert new parameters into existing
programmes; they must actually be able to rewrite the programmes.''
\cite[pp.\ 116--118]{Waddington:Paradigm}.
\end{quotation}
I therefore adopt the general idea of three different kinds of novelty
here (see below), without necessarily committing to Banzhaf \emph{et
  al}.'s specific approach. 

\begin{figure}[tbp]
\begin{center}
\includegraphics[width=0.55\linewidth]{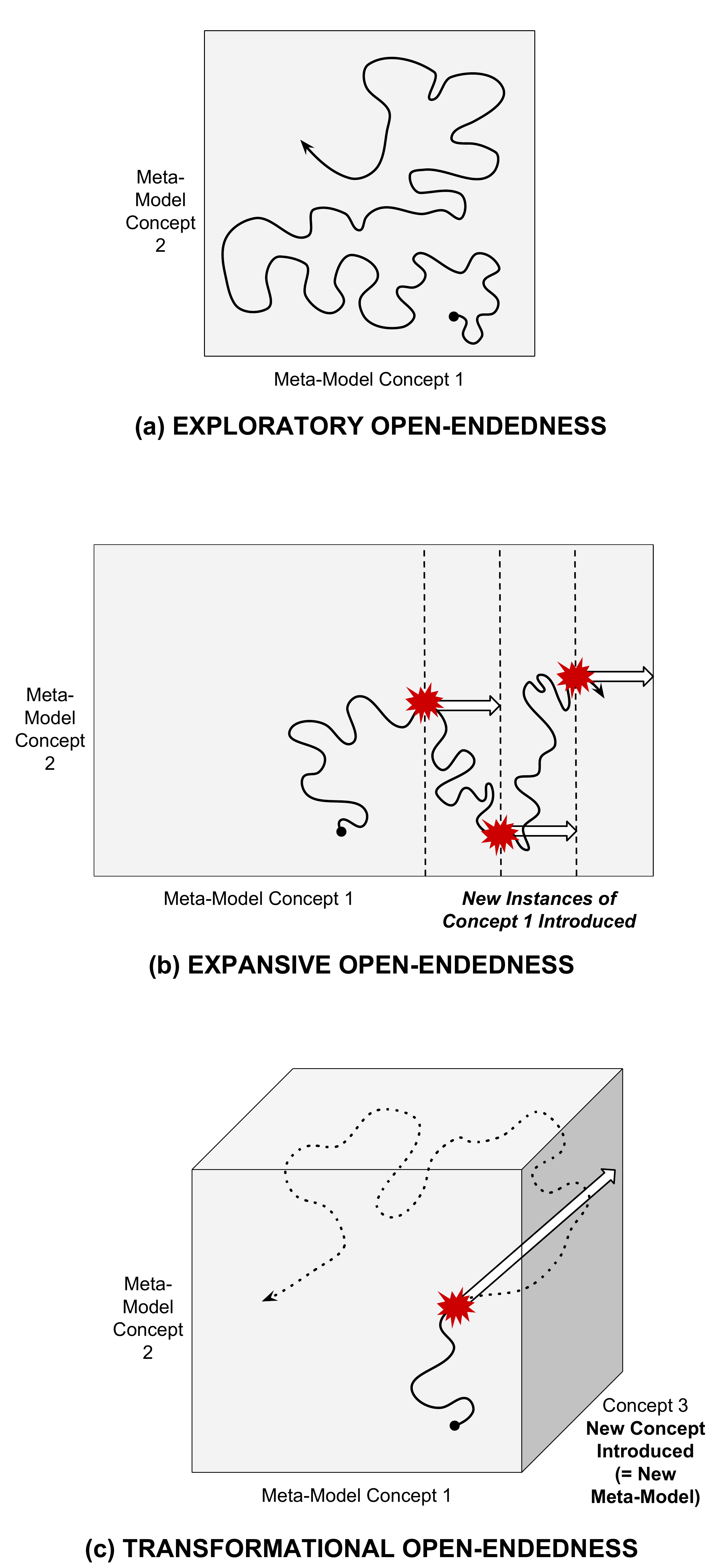}
\caption{\small{Types of open-endedness in a state space described by
    a model and its associated meta-model. See text for details.}}
\label{fig-typesOfOE}
\end{center}
\end{figure}

As discussed in previous OEE Workshops, one of the most general and
widely accepted hallmarks of \emph{open-ended evolution} is the
presence of \emph{ongoing adaptive novelty} \citep{Taylor:OEE1}. The
three different kinds of novelty therefore give rise to three
different kinds of open-endedness.
The three classes of novelty and their corresponding classes of
open-endedness are:\footnote{Banzhaf \emph{et al}.\ used the terms
  \emph{variation}, \emph{innovation} and \emph{emergence},
  respectively, in place of the terms 
  used here \cite{Banzhaf:Defining}. I have chosen to introduce new
  terminology because the existing terms (especially innovation and
  emergence) are already widely used in many different contexts and
  with many different meanings. Furthermore, the new terms nicely fit
  the concepts of open-endedness described below and illustrated in
Figure~\ref{fig-typesOfOE}. My terms fit closely with Boden's
concepts of \emph{exploratory}, \emph{combinational} and 
\emph{transformational creativity} \cite{Boden:CreativityALife}. As an
example of the potential for confusion when using Banzhaf \emph{et
  al}.'s terms, de Vladar \emph{et al}.\ have recently used the term
\emph{innovation} \cite{deVladar:GrandViews} to describe novelties 
that most closely match Banzhaf \emph{et al}.'s \emph{emergent}
novelties.}

\begin{enumerate}
\item \emph{Exploratory Novelty}: A novelty that can be described
  using the current model (e.g.\ by recombining existing components,
  or changing the values of existing parameters).

  Potential examples from biology include the production of a new 
  combination of alleles on a genome, and a change in the number of
  vertebra in a new vertebrate species. A potential example from an
  ALife evolved virtual creature system would be the appearance of a
  creature with limbs of a length that is different to what has been
  observed before. 
  
  \emph{Exploratory Open-Endedness}: The ongoing production of
  adaptive exploratory novelties.

\item \emph{Expansive Novelty}: A novelty that necessitates a change
  in the model but still using concepts present in the current
  meta-model.

  Potential examples from biology include synthesis of a
  new chemical species that has not been used in previous metabolic
  reactions, and the introduction of a new species of an existing
  genus that can occupy a new ecological niche.
  A potential example in a virtual creatures ALife system might be the
  evolution of legged locomotion where only snake-like locomotion has
  existed previously; this might represent a new instance of an
  existing meta-model concept of ``terrestrial forward locomotion''.
  
  \emph{Expansive Open-Endedness}: The ongoing production of
  adaptive expansive novelties.
  
\item \emph{Transformational Novelty}: A novelty that introduces a new
  concept, necessitating a change in the meta-model.

  Potential examples from biology include a major transition in
  individuality \cite{MaynardSmith:MajorTransitions}, the appearance
  of winged flight, and the appearance of visual sensory systems.
  A potential example in a virtual creatures ALife system might be the
  evolution of flight where only terrestrial locomotion had existed
  previously.
  
  \emph{Transformational Open-Endedness}: The ongoing production of
  adaptive transformational novelties.
  
\end{enumerate}

Note that I indicate ``potential'' examples in the list above, because
according to Banzhaf \emph{et al}.'s approach each type of novelty
is defined \emph{relative to a given model and meta-model}
\cite{Banzhaf:Defining}. 

Banzhaf \emph{et al}.\ define their three classes of
novelty in terms of the system's 
\emph{current} model and meta-model. This means, for example, that
once one major transition has been witnessed, the concept of major
transition is then added to the meta-model, so any subsequent major
transitions are not regarded as transformational.
In contrast, I suggest that novelty is defined
relative to the initial model and meta-model applied to an
evolutionary system at its inception. In that case, after a
transformational novelty appears for the first time, any further
instances of the same kind of novelty will also be labelled
transformational (and likewise for expansive novelties).\footnote{cf.\
Boden's distinction between I-creativity and H-creativity
\cite{Boden:CreativityALife}.} A defining  
feature of expansive and transformational novelties, and hence the
reason to label subsequent examples in the same class, is their
ability \emph{to open up new adjacencies in an expanded state space}
\citep{deVladar:GrandViews, Longo:NoEntailing} (see further discussion
in Section~\ref{sct-expans-trans-oe}), and this occurs each 
time such a transition arises, not just the first time.

Furthermore, Banzhaf \emph{et al}.\ chose not to classify the ongoing
production of \emph{exploratory} novelties as a type of
open-endedness. In contrast, I have chosen to do so because, even
though it takes place within a state space of fixed and finite size,
that size might well be immense. Indeed, the number of possible
combinations of entities and interactions described by a model might
easily be so astronomical that an evolutionary process could not
possibly visit all adaptive points in the space within the lifetime of
the universe. This raises the distinction between \emph{effective} OE
and \emph{theoretical} OE \citep[p.\
144]{Banzhaf:Defining};\footnote{On the distinction between
  these, Banzhaf \emph{et al}.\ say ``Should we be looking for systems
  able to continually produce open-ended events, or ‘‘simply’’ for
  systems able to produce a sufficient number of open-ended events? We
  thus distinguish systems that are theoretically open-ended from
  those that are effectively open-ended. The former may be
  demonstrable in a mathematical universe, but questionable in a
  finite universe; the latter are questionable in a  mathematical
  universe \ldots\ but may be demonstrable in a physical universe.''
  \citep[p.\ 144]{Banzhaf:Defining}.} my interest in this paper is in
effective OE. 

If we use a state space diagram to represent all possible entities and
interactions describable by a system's model and its associated
meta-model, we can represent the three different kinds of
open-endedness as shown in Figure~\ref{fig-typesOfOE}.\footnote{Any
  real system of interest will obviously 
  have far more than the two conceptual axes shown in the figure, and
  it is not clear how different instances of a concept can be mapped
  onto a scalar scale in the general case. Hence, these diagrams are
  not meant to be taken too literally, but are nevertheless useful to
  communicate an intuitive idea of the different kinds of OE.}

\subsection{Genetic and Phenotypic State Spaces}
\label{sct-g-p-spaces}

In Figure~\ref{fig-typesOfOE}, open-endedness is represented as an
ongoing traversal of the space of possible organisms. In evolutionary
systems, an organism's phenotype and behaviour are derived from a
genetic description contained in its genome. The process of generating
the phenotype from the genotype is defined by the organism's
genotype-phenotype (G-P) map. As discussed below, this map may be more
or less complex, and more or less explicit in the system's
design.

We can split the representation of phenotypic state space (P-space)
and genetic state space (G-space) into two separate diagrams.
When considering open-endedness, we are ultimately interested in
whether the system has the capacity for the ongoing production of
adaptive phenotypes in P-space. However, the ability of an
evolutionary system to explore P-space is fundamentally affected by
the nature of the G-P map as the evolutionary processes of
reproduction and variation of the genome explore different points in
the genetic state space (G-space). 

\begin{figure}[tb]
\begin{center}
\includegraphics[width=0.65\linewidth]{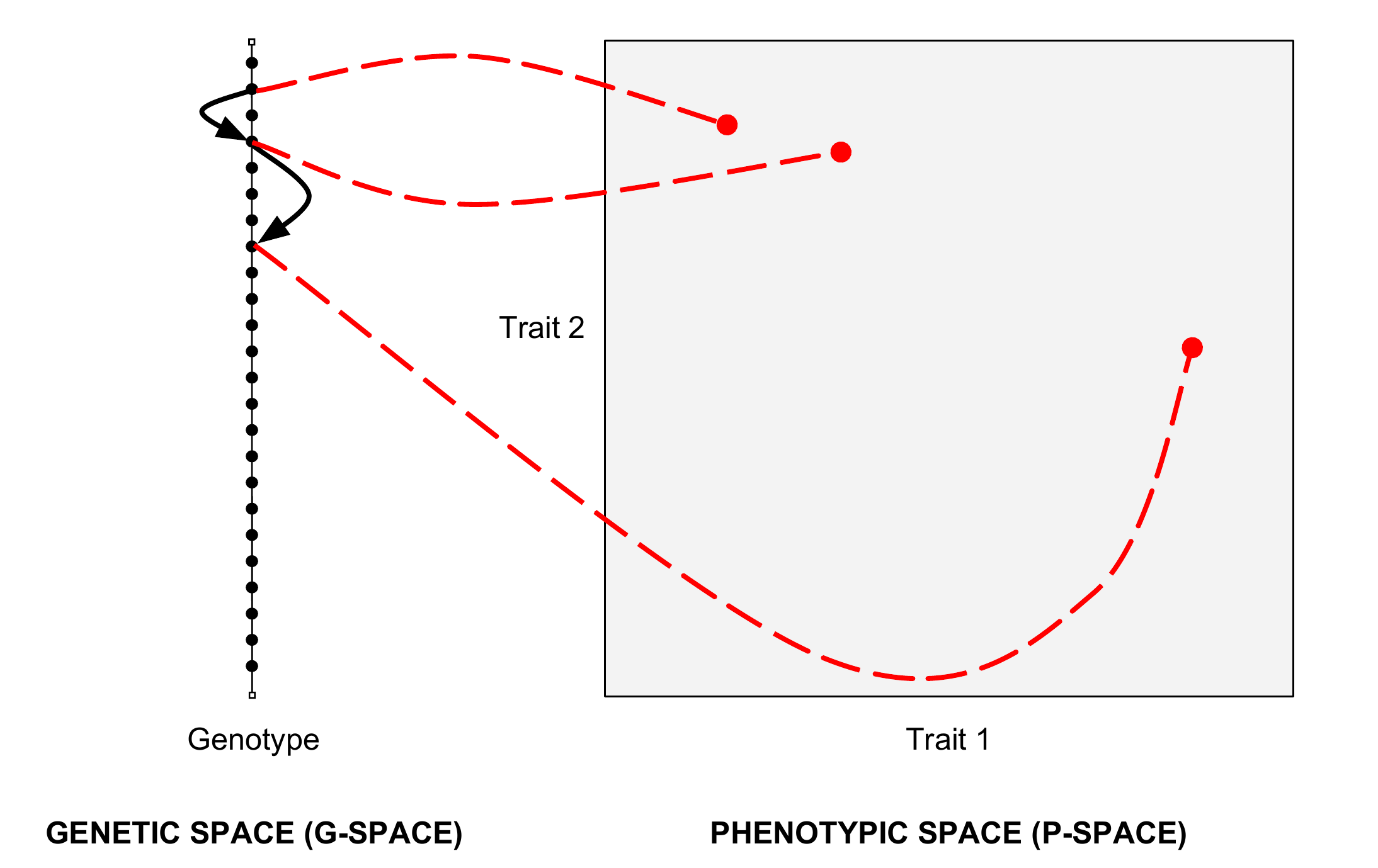}
\caption{\small{Genetic Space and its relation to Phenotypic
    Space. The mapping from G-space to P-space is defined by the G-P map.
    The mapping might be such that small moves in G-space can
    sometimes result in large moves in P-space. Note that the
    dimensionality of G-space might be different to that of P-space.}}
\label{fig-gspace}
\end{center}
\end{figure}

A simple example of G-space, along with its relationship to P-space,
is shown in Figure~\ref{fig-gspace}.
Note that the dimensionality of the G-space is not necessarily the
same as that of the P-space: the relationship is determined by the G-P
map, which can be of arbitrary form and can also depend upon the
system's global laws of dynamics and the local context in which the
phenotype is generated.\footnote{In terms of the formalism to be
  introduced later (Section~\ref{sct-generation}), the G-P map is the
  $M_L$ function in Equation~\ref{eqn-generation}, the global laws of
  dynamics are represented by the $L$ subscript of that function, and
  the local context is represented by the function's $c_a$ and $c_b$
  parameters.} 

%
As shown in Figure~\ref{fig-gspace}, small
moves in G-space might result in large moves in P-space, depending
upon the nature of the G-P map. Furthermore, different mappings
will lead to different paths in P-space for a given set of moves in
G-space. Hence, the nature of the mapping, and whether or not the mapping
itself has the ability of evolve over time, will fundamentally affect
the system's ability to explore different areas of P-space.

\subsubsection{Relationship between G-space and open-endedness in P-space}

In some cases, a system might exhibit effective transformational
open-endedness in P-space even with a fixed G-space; 
in Section~\ref{sct-accessnewstates} we will see how this might
happen.\footnote{As we will see in Section~\ref{sct-accessnewstates},
  this can come about where there is a \emph{non-additive
    compositional complexity} in the building blocks of the phenotype,
  or the presence of a \emph{transdomain bridge}.}
%
While effective transformational open-endedness is possible in a fixed
G-space, one might think that a more obvious way to achieve it is to
allow the number of genes on the genome to grow---leading to an
expanding G-space. If the size of the genome can potentially expand
without limit, we have what is referred to in the evolutionary biology
literature as an \emph{indefinite hereditary replicator}
\citep{MaynardSmith:MajorTransitions}. 
All else being equal, a larger genome can (but does
not necessarily) specify a more complicated phenotype. While this can
indeed be the case,
the ability of a larger genome to specify expansive or
transformational novelties
depends upon the capacity of the
additional genes to specify new traits. This can be achieved (as in
the fixed G-space case) through the methods to be discussed in
Section~\ref{sct-accessnewstates}.

\section{Evolutionary Processes}
\label{sct-formalism}

Having introduced some concepts and definitions relating to open-ended
evolution in the previous section, I now discuss some high-level
general features of evolutionary systems, and introduce a formalism to
describe them. This will then provide a framework that can be used to
explore different ways in which open-endedness may be introduced into
an evolutionary system, which we will do in
Section~\ref{sct-routes}. I do not claim that the formalism presented
in this section if particularly novel or of wide applicability beyond
the current discussion; its main purpose is to emphasize various
parameters and routes of interaction between processes that are not
usually explicitly denoted. 
Within the context of this paper
it provides a useful tool for considering various interactions of
interest in an evolutionary system, and how they relate to
open-endedness. 

Considering evolutionary systems in general---including, for example,
biological evolution, genetic algorithms, evolutionary robotics
systems, and systems of self-reproducing computer code---we can
discern three fundamental processes that any such system must
instantiate in some form or other:\footnote{The three core processes
  of a Darwinian evolutionary process are often stated as \emph{variation},
  \emph{differential reproduction}, and \emph{inheritance}. This
  ignores the process of \emph{generation} (of phenotypic behaviour from
  genetic description) stated in the list given here, which is
  important in the current context. On the other hand, we collapse the
  processes of \emph{variation} and \emph{inheritance} into a single process
  ``\emph{reproduction with variation}'' as a simplifying step in this
  discussion. This is a valid simplification as long as we are only dealing
  with systems where the main source of variation among individuals
  arises during the reproduction of a parent(s) to generate an offspring.}

\begin{enumerate}
\item The \emph{generation} of the phenotypic behaviour of an individual
  from its informational (genetic) description.
\item The \emph{evaluation} of phenotypes to determine which ones get
  to reproduce. In its most general form the evaluation also
  determines the schedule of reproduction (rate and number of
  offspring) and lifetime of the individual.
\item The \emph{reproduction with variation} of successful
  individuals.
\end{enumerate}

The explicitness and complexity of implementation of each of these
processes varies significantly from one type of system to another.
In some cases a process might be implemented \emph{extrinsically} as
a special purpose hard-coded mechanism acting upon the system, whereas
in other cases the process might be provided \emph{intrinsically} by a
mechanism implemented within the system itself.
Intrinsic mechanisms may rely exclusively upon the general laws of
dynamics of the system (e.g.\ a simple self-replicating molecule), or
they may be under sophisticated evolved control provided by
the organisms themselves (e.g.\ the generation and reproduction
mechanisms of modern biological organisms).

In some cases it may be easy to
overlook the presence of a particular process; for example, in systems
such as Tierra \citep{Ray:Approach} and Avida \citep{Ofria:Avida}, one
might think there is no process of generation from genotype to
phenotype, but a closer look shows that the phenotype comes about
through the action of the system's (virtual) CPU that executes the
instructions present in a program's genotype
\citep{Taylor:Creativity:Book}. One way or another, these three
processes are implemented by \emph{all} evolutionary systems. 

\begin{figure}[tb]
\begin{center}
\includegraphics[width=0.5\linewidth]{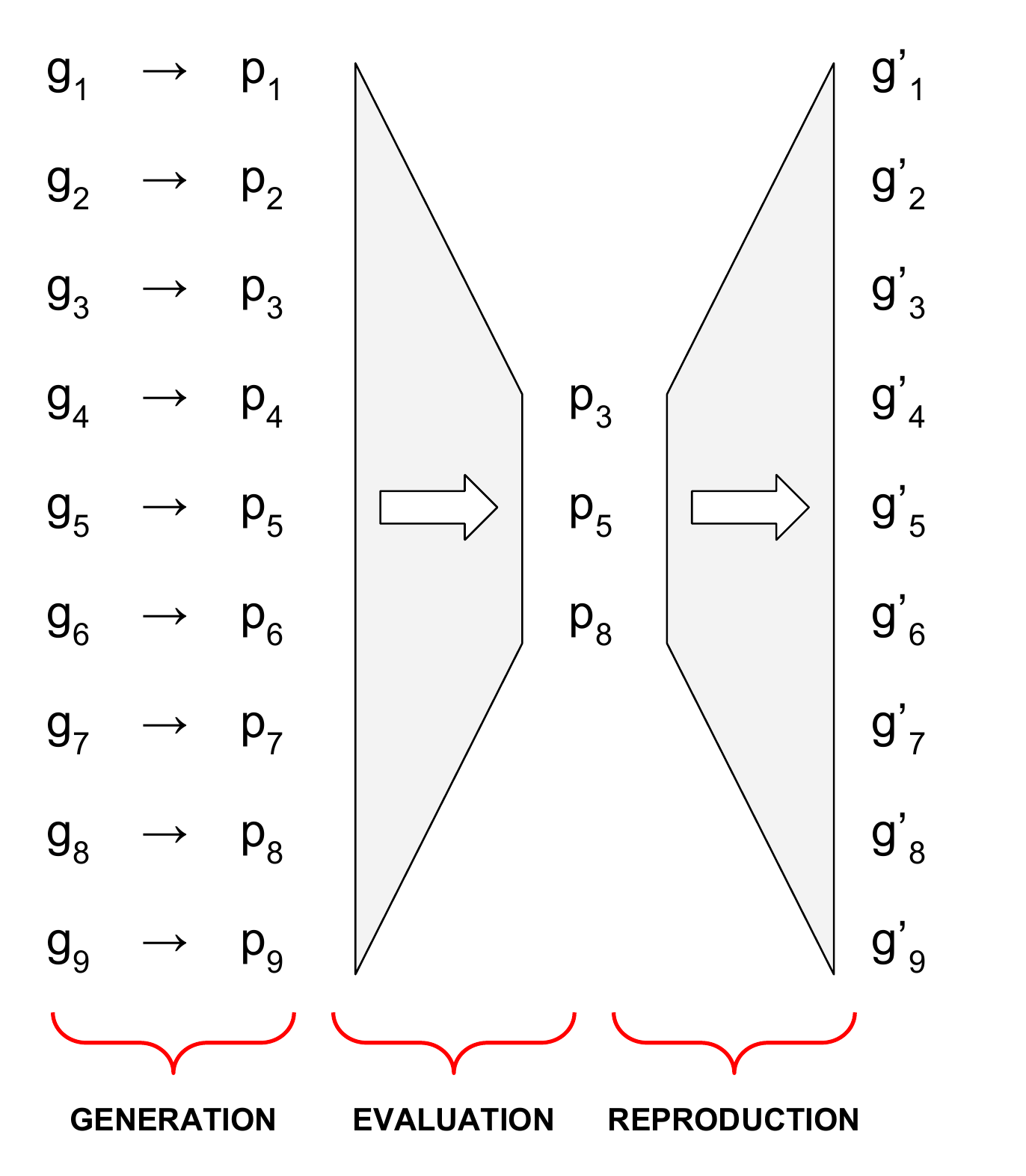}
\caption{\small{Schematic overview of key processes that must be implemented
  by any evolutionary system. Note that the timing and duration of
  each process does not necessarily need to be the same for each
  organism in the system, and the total number of individuals does not
  necessarily need to be constant from one generation to the next.}}
\label{fig-processes}
\end{center}
\end{figure}

A schematic overview of how the three processes act upon a population
of individuals in shows in Figure~\ref{fig-processes}. Each process is
explained in more detail below, and a formalism is introduced to make
explicit various aspects of each process and interrelationships
between the processes.\footnote{The formalism is by no means complete,
  but it does at least emphasize influences on, and
  interrelationships between, the three processes. Some weaknesses of
  the formalism are discussed at the end of the paper, along with
  suggestions for future improvements.}

\subsection{Generation}
\label{sct-generation}

The process of generation can be represented in a very general form as
follows: 

\begin{eqnarray}\label{eqn-generation}
  p = M_L(g, c_a, c_b)
\end{eqnarray}

where $g$ is the genotype, $p$ is the resulting phenotype, $M$ is the
function that generates $p$ from $g$, i.e.\ the genotype-phenotype 
(G-P) map, $L$ indicates fixed global laws (``laws of physics'')
acting upon the system (which may contribute to determining the
outcome of the generation process, e.g.\ self-organisational processes
arising from laws of physics and chemistry in the biosphere, or the 
CPU interpretation of instructions in Tierra), $c_a$ indicates the
local abiotic context (environmental conditions) in which the
generation process occurs, and $c_b$ indicates the local biotic
context (influence of other organisms on the process).

\subsection{Evaluation}

The evaluation of a phenotype to determine its evolutionary
significance, i.e.\ if and when it reproduces, how many offspring it
produces, whom it mates with, and how long it lives, can be represented
very generally as follows:

\begin{eqnarray}\label{eqn-evaluation}
  (l, \overline{s_r}, \overline{p_m}) = E_L(p, c_a, c_b)
\end{eqnarray}

where $E$ is the evaluation function, $L$ indicates fixed global laws
acting upon the system (which may contribute to determining the
outcome of the evaluation, e.g.\ laws of aerodynamics determining the
ability of a bird to fly), $p$ is the phenotype, $c_a$ and
$c_b$ are the local abiotic and biotic context (as above), $l$ is the
resultant lifetime of the phenotype as determined by the evaluation process,
$\overline{s_r}$ is a vector representing the phenotype's resultant
\emph{reproduction schedule} (i.e.\ the number and timing of applications of
the reproduction process on the individual), and $\overline{p_m}$ is a
vector representing the phenotype's resultant \emph{mate set}, i.e.\ the
mate(s) that will participate in the individual's reproduction process
(in the most general case, this set may be empty or of any non-empty
size). 

\subsection{Reproduction with Variation}

Finally, the reproduction process can be represented in general form
as follows:

\begin{eqnarray}\label{eqn-reproduction}
  g' = R_L^{\overline{s_r}}(p, \overline{p_m})
\end{eqnarray}

where $R$ is the reproduction function, $L$ indicates fixed global laws
acting upon the system (which may contribute to determining the
outcome of the reproduction process, e.g.\ by specifying global
mutation rates), $p$ is the phenotype, $\overline{p_m}$ is the
mate set as determined by the evaluation process, $\overline{s_r}$
is the reproduction schedule as determined by the evaluation process,
and $g'$ is the resultant new genotype.\footnote{In
  Equation~\ref{eqn-reproduction}, $\overline{s_r}$ is presented as a
  superscript to indicate that it determines \emph{when} $R$ is applied.}

The reproduction function may incorporate any of a variety of
different procedures depending upon the evolutionary system under
consideration, including mutations of various kinds, recombination,
gross chromosomal rearrangements (GCRs), error correction mechanisms,
and so on.

Note that $R$ is stated as a function of $p$ rather than $g$. It is
assumed here that $p$ has access to the original $g$ that created it,
so that $R$ could produce the new $g'$ by simply copying $g$. But
using $p$ in the function allows for a more general representation
that can also describe the transmission of acquired characteristics
from $p$ to $g'$ (Lamarckian evolution) if relevant.

\section{Routes to Achieving Open-Endedness}
\label{sct-routes}

Having covered the three different kinds of open-ended evolution and a
general formalism with which to describe the key processes of an
evolutionary system, I now
use
the formalism
to
identify various routes by which open-endedness can be introduced in
the design of an evolutionary system.
As the formalism encapsulates a fairly standard Darwinian view of
evolution (with extra emphasis on inputs and parameters to processes),
the idea is to see how far this view will get us in explaining
how open-endedness may arise in an evolutionary system.

\begin{figure*}[tb]
\begin{center}
\includegraphics[width=1.0\textwidth]{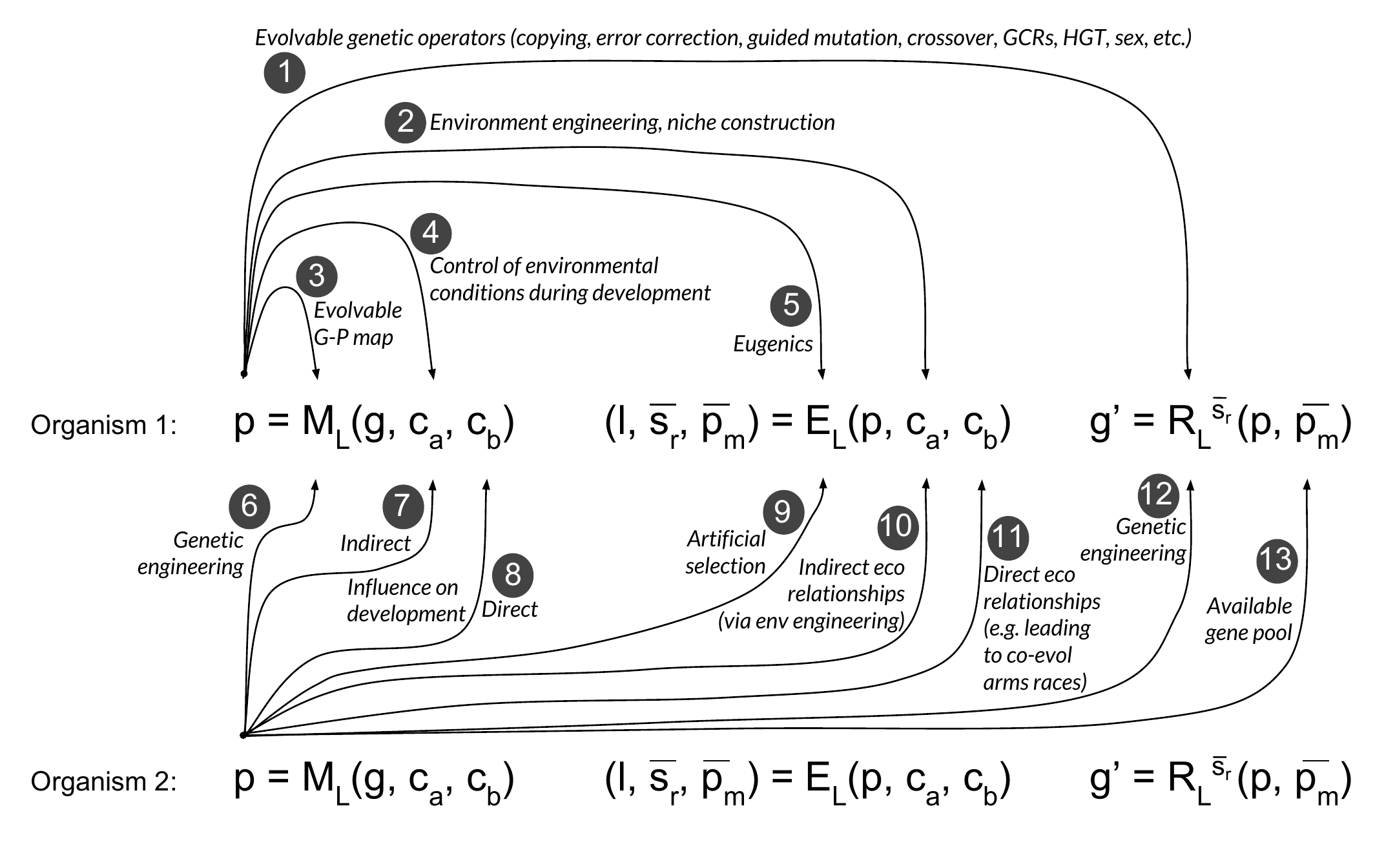}
\caption{\small{Potential routes to exploratory open-endedness in an evolutionary system.} }
\label{fig-routes}
\end{center}
\end{figure*}

In a simple system of non-interacting individuals that reproduce with 
variation according to static evaluation and 
reproduction functions, the individuals will evolve towards a local
optimum in the adaptive landscape. At this point, stasis (or at least
quasi-stability) will be reached.
We are therefore looking for routes by which the organisms ($p$ in the
equations) can influence the three evolutionary
processes described in the formalism, to enable them to escape from
this situation and maintain an ongoing exploration of P-space.
These could involve an organism
causing a change in the implementation of one of the functions defined
in Equations~\ref{eqn-generation}--\ref{eqn-reproduction}, or causing
a change to one of the parameters of those equations.
All such routes are mapped out in Figure~\ref{fig-routes}; the top
half of the figure (the line labelled ``Organism~1'' and the arrows
above it) shows how an organism $p$ can affect the evolutionary
processes associated with its own lineage, and the bottom half
(the line labelled ``Organism~2'' and the arrows above it) shows how
other organisms can affect the parameters of the evolutionary
processes of Organism 1.\footnote{The figure does not show an arrow
  from Organism 1 $p$ to the $c_b$ parameters of $M_L$ and $E_L$
  because $c_b$ is, by definition, the context caused by \emph{other}
  organisms.}
Each of these routes is
discussed in more detail in Section~\ref{sct-explore-oe}.

As revealed in the following discussion, an analysis of open-endedness
based upon the formalism only really addresses issues concerning
\emph{exploratory} open-endedness. This illustrates why traditional
approaches to modelling evolutionary systems based upon the processes
of generation, evaluation and reproduction with variation do not
provide much insight into the more interesting kinds of open-endedness,
i.e.\ expansive and transformational open-endedness. In the following
discussion I also suggest routes by which these other two types of
open-endedness can be achieved, although these are more tentative
suggestions offered without the support of the formalism.

Before discussing the different routes specifically, I begin with some
general comments on the distinction between intrinsic and extrinsic
implementations of the evolutionary processes.

\subsection{Intrinsic and Extrinsic Implementations}
A cross-cutting issue in the quest for open-endedness described in the
following discussion is the extent to which each of the specific
processes is defined \emph{intrinsically} within the system by being
\emph{implemented through the components and dynamics of the system
  itself}. In contrast, all existing artificial evolutionary systems
define some or most of these processes \emph{extrinsically} to the
evolving system as a special purpose hard-coded mechanism. Banzhaf
\emph{et al}.\ refer to extrinsically implemented mechanisms as
\emph{shortcuts} \cite[p.\ 146]{Banzhaf:Defining}. 

The importance of using an intrinsic evaluation process in
computational models of biological evolution has been recognised for a
long time (e.g.\ \citep{Packard:Intrinsic}) and indeed was a feature of some
of the earliest implementations of computational evolutionary systems
(e.g.\ \citep{Barricelli:SymbiogeneticEvolution}, \cite{Conrad:Evolution}).
Here I consider the benefits of implicit implementations not just of
the evaluation process, but also of the generation and reproduction
processes. The key benefit of processes instantiated intrinsically by being
explicitly implemented within the system itself is that it allows the
possibility that the implementation---the process---can itself
change. This opens the door for the G-P mapping, the evaluation processes
and the reproduction and variation processes to evolve as the system
unfolds.

While it is possible to imagine extrinsically coding not
just a process but also mechanisms for changing the process, such a
process would still only be able to change and evolve in the
hard-coded ways provided by the extrinsically defined change
mechanism. In contrast, for intrinsically implemented processes, not 
only might the process evolve, but \emph{the evolvability of the
  process} might itself evolve.

\subsection{Exploratory Open-Endedness}
\label{sct-explore-oe}

As stated above, in a simple system of non-interacting individuals
that reproduce with variation according to static evaluation and 
reproduction functions, the individuals will evolve towards a local
optimum in the adaptive landscape beyond which the variational methods
of the reproduction function can no longer take them to a state with
higher fitness.

To introduce the possibility of ongoing adaptive novelty we therefore
need mechanisms that can ``shake up'' the system. This can be achieved
by allowing for intrinsic means for ongoing modification of the
adaptive landscape experienced by an individual (i.e.\ changes
relating to the evaluation function $E_L$), of the topology of genetic
space (i.e.\ changes relating to the reproduction and variation
function $R_L$), or of the mapping between genotype and phenotype
(i.e.\ changes relating to the generation function $M_L$).
As explained above, the routes for implementing these kinds of
mechanisms within our formalism are identified in
Figure~\ref{fig-routes}. We now look at each of these in more detail.

If the reproduction and variation processes ($R_L$) are themselves
implemented intrinsically by the individual organisms (Route 1 in
Figure~\ref{fig-routes}), the individuals might be able to jump out of
the local optimum by bringing new areas of G-space and P-space into
reach of the variational operators. Route 1 represents topics in the
literature concerning \emph{the evolution of evolvability} (evo-evo),
such as evolvable genetic operators, including copying processes,
error correction, mutator genes for guided mutations, crossover
mechanisms, gross chromosomal rearrangements, horizontal gene
transfer, etc.\
(e.g.\ \cite{Pigliucci:EvoEvo}, \cite{Crombach:Evolution}). 

Another route to improving evolvability is
provided by Route 3; allowing the G-P map ($M_L$) to evolve by having
it intrinsically implemented by the individual organisms. As
discussed earlier (see Figure~\ref{fig-gspace}), the nature of the G-P
map dictates which regions of P-space can be easily explored.
Implementing the G-P map intrinsically potentially allows it to
evolve such that mutations are more likely to produce adaptive
variations in P-space. Route 3 represents topics in the literature
such as evo-devo, facilitated variation and developmental robustness
(e.g.\ \cite{Nuno:Computing}, \cite{Gerhart:Theory}).

An alternative possibility for an organism to influence the result of
the G-P mapping process is for it to exert some control over the
environmental conditions under which the process occurs (in the case
where the mapping is sensitive to such conditions); this is
represented by Route 4 in Figure~\ref{fig-routes}. However, in
biological systems this is more likely to be a route 
whereby organisms \emph{reduce} variability (e.g.\ canalization
\cite{Waddington:Canalization}) rather than increase it.

While changes to $M_L$ and $R_L$ via Routes 1 and 3 (evo-evo and
evo-devo) might be sufficient to prevent the system becoming stuck in
a local optimum state, they are not in themselves sufficient to
achieve an ongoing exploration of P-space, as the system will still
halt when all individuals have reached a statically-defined optimum
fitness.
In order to provide an ongoing drive for exploring P-space, ongoing
change in the adaptive landscape is required. This can be achieved by
inducing ongoing changes to an individual's evaluation function
$E_L$, which can be realised via Routes 2, 5, 9, 10 or 11.

Routes 2, 10 and 11
all change the context in which
the individual is evaluated: Routes 2 and 10 change the abiotic
context, and Route 11 changes the biotic context. In Route 2 the change
is brought about by the focal individual itself (e.g.\ by
environmental engineering \cite{Jones:Positive}, or niche construction
over longer timescales \cite{OdlingSmee:Niche}), whereas in Route 10 it
is brought about by the other individuals that influence the
evaluation function (e.g.\ environmental engineering by other
species). Route 11 represents direct ecological relationships, leading
to processes such as co-evolutionary arms races \cite{VanValen:New}.

Beyond changing the context in which evaluation occurs, it might also
be possible to intrinsically change the evaluation function ($E_L$)
itself (i.e.\ to change the fundamental factors determining an
individual's longevity and fecundity). This might arise when
an intelligent species that practices artificial selection (e.g.\
farming or eugenics) evolves within the system. These mechanisms are
represented by Routes 5 and 9.\footnote{Note that this case of an
  intrinsically-defined evaluation function is different to the case
  of evolutionary algorithms that explicitly avoid imposing
  notions of objective fitness, such as \emph{Minimal Criterion Novelty Search}
  \cite{Lehman:Revising}. These algorithms still have an
  extrinsic evaluation function at their core (in the case of Novelty
  Search, this includes the method by which the novelty of a new
  organism is calculated).}

Regarding the other two functions ($M_L$ and $R_L$), it would be very
unusual to have a system where one organism could directly change
another organism's implementation of these functions. However, this is
at least conceivable, and perhaps the human species is approaching
that ability with advances in genetic engineering (Routes 6 and 12).

Additional routes through which other individuals in the system might
promote ongoing exploration of P-space include Routes 7, 8 and
13. Routes 7 and 8 are processes whereby the production of a
phenotype from a genotype is affected by the local context: Route 7
represents the local abiotic environment and Route 8 the local biotic
environment. Neighbouring individuals can be involved in Route 7 as
well as Route 8, through the processes of environment engineering and
niche construction. Route 13 represents the available gene pool
provided by potential mates in the local context, i.e.\ the raw
material upon which the genetic operators might act in Route 1.

The role of all of the processes involving interactions with other
individuals via the parameters of the functions (Routes 7, 8,
10, 11 and 13)
can be boosted if the local context experienced by an individual and
its descendants changes over time. An obvious route for achieving this
is through the provision of a spatial environment and means by which
individuals can move (actively or passively) around the environment. 

It should also be recognised that the formalism developed here is not
exhaustive, as it concentrates only on processes that affect
individuals. It does not explicitly deal with population-level effects
that are also relevant in promoting ongoing exploration of G-space and
P-space. Topics from the evolutionary population dynamics literature
such as finite sampling, drift, adaptive radiations, and neutral
networks, are additional mechanisms by which the ongoing exploration
of the adaptive landscape might be promoted.

To summarise the preceding discussion, the routes to open-endedness
depicted in Figure~\ref{fig-routes} represent mechanisms by which
organisms can promote ongoing evolutionary activity by modifying the
adaptive landscape, the topology of genetic space, or the nature of
the G-P map. 

However, while these routes promote ongoing activity within a given
P-space, none of them cause the expansion of P-space itself. In other
words, the routes to open-endedness suggested by a fairly standard
Darwinian analysis, as represented by the formalism, relate to
\emph{exploratory} open-endedness only; they do not directly help us
in our search for \emph{expansive} or \emph{transformational}
novelties.

\subsection{Expansive and Transformational Open-Endedness}
\label{sct-expans-trans-oe}

We will now discuss concepts not explicitly
covered by the formalism that may be required to produce the other
kinds of open-endedness involving expansive and transformational novelties.
These more interesting kinds both involve the
discovery of \emph{door-opening}\footnote{To borrow a term from Bedau
  \citep{Taylor:OEE1}.} states in P-space that open up an expanded
space of new adjacencies, as exemplified by the red flashes in
Figures~\ref{fig-typesOfOE}(b) and \ref{fig-typesOfOE}(c).

There are various issues involved in how these might come about in an
evolutionary system.
The following discussion address two of the most important questions:
\begin{enumerate}
\item Where does this extra space of possibilities come from?
\item How can the evolutionary system access the new states via
  intrinsic mechanisms? 
\end{enumerate}

\subsubsection{(Q1) Expanding the state space}

Regarding Question 1, in the biological world the answer is that the
\emph{extra space was always there} in the complexity of the laws of physics
and chemistry---it is just a matter of biological systems evolving to
make use of the existing complexity (by methods pertaining to 
Question 2). Engineered \emph{physical} evolutionary systems can also
make use of this existing complexity---indeed, the most impressive
instances of transformational novelties arising in artificial systems
are found in physical systems, e.g.\
\citep{Cariani:EvolveEar,Bird:EvolvedRadio}.

This situation exemplifies the fact that our definitions of novelty
and open-endedness, as presented in Section~\ref{sct-novelties}, are defined
\emph{relative to our model} (and meta-model) of the system. The state
space of the actual system has not expanded, but an expansive or
transformational novelty reveals a \emph{deficiency in the model} of the
system regarding its ability to describe the actual system. These
kinds of novelties therefore require an expansion of the model.

In the case of computational evolutionary systems, the same solution
of providing a world with rich possibilities for complexity in
its laws of dynamics and interactions is also an option. It is notable that most
existing ALife work with computational evolution takes place in very
impoverished virtual environments. But there is also another
possibility with computational systems: to dynamically increase
P-space as the system unfolds. One route by which this might be
achieved would be to open up the system by allowing it to access
additional resources on the internet (e.g.\ stock trading agents with
the ability to discover and utilise new online data sources to improve
their performance).\footnote{This idea has been
  discussed by Boden among others \cite{Boden:CreativityALife}. See
  \citep{Taylor:WebAL} for many pointers to how this might be
  implemented.}

\subsubsection{(Q2) Accessing new states}
\label{sct-accessnewstates}
Consideration of the mechanisms involved in biological evolution
suggests at least two general ways in which Question 2 can be 
addressed: 

\paragraph{(a) Domains, exaptations and transdomain bridges}
Components in physical systems possess multiple properties in
different domains (e.g.\ mechanical, chemical, electrical,
responsiveness to electromagnetism, pressure, etc.).
Indeed, the distinction between an expansive and transformational
novelty can by viewed as the difference between \emph{a door-opening
  novelty in the same domain} versus \emph{a door-opening novelty in a
  different domain}, respectively. In this view, the distinction
between expansive and transformational novelty depends upon an
observer's ontology of domains; this is a more specific
interpretation of the picture of models and meta-models
introduced earlier. 

A common mode by which innovations arise is \emph{exaptation}, where a
structure originally selected for its properties in one domain
coincidentally has adaptive properties in a different domain which
then become a new focus of selection \citep{Gould:Exaptation}. In this
situation, the multi-property component has acted as a
\emph{transdomain bridge} to open up a new domain 
for potential exploitation by the organism---this would represent a
transformational novelty. This mechanism can also produce expansive
novelties if the components have multiple properties within the same
domain, e.g.\ multifunctional enzymes
\citep{Kacser:Evolution}.\footnote{The importance of multifunctional
  components for biological evolvability and robustness has been
  argued by various authors, e.g.\ \citep{Goldenfeld:Life,
    Whitacre:Degeneracy}.} The latter case can be labelled a
\emph{intradomain bridge}.

Another example, provided by Dawkins, is the evolutionary appearance
of segmented body plans in animals \cite{Dawkins:EvoEvo}. 
While the first segmented animal
might have been unremarkable in terms of its functionality, and just
an exploratory novelty, it gave rise to a radiation leading to a whole
new phyla with \emph{new possibilities for behaviour} (i.e.\ expansive and/or
transformational novelties),
such as new
possibilities for locomotion arising from the free movement afforded
to organisms with a segmented spinal cord. In close alignment with our
terminology of \emph{door-opening novelties}, Dawkins describes
discoveries of this kind as ``watershed events \ldots\ that open
floodgates to future evolution'' \citep[p.\
218]{Dawkins:EvoEvo}.\footnote{The distinction between the exploratory
  discovery of a door-opening state and the 
  potential it introduces for expansive or transformational novelties
  in function is similar to G.\ Wagner's distinction
  between (his conception of) \emph{novelties} and
  \emph{innovations} \cite{Wagner:Evolutionary}. We further discuss
  this distinction at the end of the paper.}

Most computational evolutionary systems lack significantly
multi-property components, and therefore miss out on this route to
transformational novelty. The examples that currently benefit the most from
this route to novelty are those in which the evolving agents are
embedded within a simulated physical environment \cite{Taylor:Recent}.

\paragraph{(b) Non-additive compositional systems}
An alternative route for accessing new states hinges on the mechanism
by which a phenotype is generated from a genotype ($M_L$). To take a
very general view, we can see this process as the construction of a
structure and/or behaviour by the specific arrangement of a number of
components drawn from a given set of component types.
I will call this mode of construction a \emph{compositional
  system}.\footnote{I use the term 
  \emph{compositional} rather than Boden's \cite{Boden:CreativityALife}
  \emph{combinational} to emphasise that the size of structures may
  increase, and that the specific arrangement and connections between
  components might be important.}
Examples from biology range from the construction of a protein from
amino acids drawn from a set of 20 different types, to the
construction of an termite colony from termites drawn from a set of
different castes. Examples from ALife include the construction of a
neural network controller from a given number of neurons and
connections. In many cases, particularly in biology, there may be
hierarchical levels of composition; see \citep{Banzhaf:Defining} for
an extensive discussion of levels and hierarchies.

Compositional systems can arise in many different domains, such as
chemistry, physics, and information systems. To a first degree of
approximation, we could view prokaryotic life as an exploration of
compositional \emph{chemistry} and multicellular eukaryotic life as an
exploration of compositional \emph{physics}.\footnote{This is obviously a
  gross simplification, as all domains of life utilise both chemistry
  and physics.} Furthermore, animals with nervous systems, and
ALife agents with evolved controllers, engage in the
exploration of compositional \emph{information} systems.

Note that the ability of a lineage to concurrently explore multiple
compositional domains is in itself an enabler of exploratory
open-endedness, as it can prevent evolution from getting stuck in a
local optima in any one domain by providing an  \emph{extradimensional
  bypass}.\footnote{This concept was introduced by
  Conrad \cite{Conrad:Geometry} and later named by Gavrilets
  \cite{Gavrilets:Dynamical}.}

We can distinguish between \emph{additive compositional systems} and
\emph{non-additive compositional systems}. For \emph{additive}
systems, the functionality of the resulting product is an
amplification of the existing function of the components (e.g.\
joining a number of batteries in serial to create a new battery with a
greater voltage). For \emph{non-additive} systems, the act of
composition can introduce new functionality depending upon the
specific arrangement and connections between the parts (e.g.\
composing a computer algorithm out of a specific set of subroutines
and individual instructions). In some non-additive compositional
systems such as biomolecular chemistry, this can also be a route to
accessing new domains (e.g.\ as is the case with the production of a
photoreceptor protein such as rhodopsin from its amino acid
sequence).\footnote{It is also possible that a system might have a
  mixture of different types of components, some of which are additive
  and others non-additive.}

While additive compositional systems result only in exploratory
novelty, they can play an important role in enabling later expansive
or transformational novelties. To take the battery example mentioned
above, the creation of a new battery with higher voltage does not
introduce new functionality in itself, but the higher voltage might
make other processes and reactions possible that were not previously
achievable. Another example is the previously discussed case of the
evolutionary appearance of segmented body plans in animals.
So additive compositional systems can create door-opening
states that subsequently lead to expansive or transformational
novelties.  

In contrast, \emph{non-additive} compositional systems can lead
\emph{directly} to expansive or transformational novelties. For
example, building proteins from amino acids can produce new molecules
possessing \emph{expansive} novelties in the chemical reaction repertoire:
``Once a new molecule appears for the first time in the chemosphere
new interactions and further adjacencies emerge'' \citep[p.\
4]{deVladar:GrandViews}. As already mentioned above, protein
building can also potentially lead to \emph{transformational}
novelties, as might be the case with the production of a
photosensitive rhodopsin molecule. 

I close this section with a few remarks about compositional systems
and the evolution of complexity. While the complexity of organisms and
interactions does not necessarily increase in an evolutionary system,
those that employ compositional systems in the production of
phenotypes have a clear capacity for \emph{cumulative compositional
  complexity} as evolution builds upon what has gone before. 
This capacity would appear to be particularly pronounced in
non-additive compositional systems, where new compositions can offer
direct routes to expansive and transformational novelty.
Furthermore, compositional systems
able to cumulatively produce hierarchical organisations are
particularly suitable as a basis for the evolution of complexity
\citep{Simon:Architecture}.
Increases in complexity in these cases will
be aided by the usual drivers of complexity discussed in the
evolutionary biology literature, such as co-evolutionary arms races
\cite{VanValen:New}.

\section{Final remarks}
\label{sct-final-remarks}

The framework presented above can act as a guide for categorising and
comparing the OE potential of existing systems. For example, von
Neumann's CA implementation of a self-reproducing system concentrated
heavily on the role of the laws of dynamics ($L$) in its intrinsic
implementation of the evaluation function $E_L(p, c_a, c_b)$, but
ignored the organism's local context ($c_a$ and $c_b$), making the
system very brittle to perturbations.
Tierra implements $E_L$ intrinsically, but $M_L$ is extrinsic and
trivial, and the abiotic environment (as represented by the laws of
dynamics, $L$) is very impoverished. Geb \citep{Channon:Unbounded}
features intrinsic $E_L$ applied to non-additive compositional
controllers (neural networks), but implements $M_L$ and $R_L$
extrinsically.
Most implementations of Novelty Search
\citep{Lehman:Abandoning} implement two or all
three key processes ($M_L$, $E_L$ and $R_L$) extrinsically---although
in many cases this is applied to non-additive compositional
controllers and other compositional systems.
A comprehensive examination of existing systems along these lines
would provide clear indications of how the OE potential of future
systems could be improved. 

As demonstrated in the preceding discussion, the framework can act as
a map of the territory of open-endedness. This is useful for showing
how the diverse body of relevant existing theory fits into the overall
picture, in addition to aiding the categorisation and comparison of
systems as outlined above.
The discussion has revealed that considerations of generation,
evaluation and reproduction with variation indicate routes to
\emph{exploratory} open-endedness only; in order to understand the more
interesting cases of \emph{expansive} and \emph{transformational}
open-endedness, we need to consider not the properties traditionally
studied by population genetics, but rather the nature of the building
blocks out of which individual organisms are constructed, and the laws
and properties of the environment in which they exist. 

The presented framework makes explicit various influences and
interrelationships between the fundamental processes required for
evolution. Nevertheless, as stated above, there are clearly areas
where the framework could be further improved. It is currently weak at
representing important processes occurring above the level of the
individual organism. 
For example, Figure~\ref{fig-routes} does not currently
capture evolutionary population dynamics concepts such as finite
sampling, drift, neutral networks, and so on. 
The framework could certainly be expanded to be more
explicit about such population-level effects.

The current work might also benefit from a more sophisticated
treatment of the concept of \emph{behaviour}.
One approach would be to replace the current use of
G-space and P-space with a threefold distinction between Parameter
Space, Organisation Space and Action Space. Items in Parameter Space are
informational specifications of particular Organisations
(configurations of material) 
in Organisation Space. These organisations are situated in a specific
environment that provides boundary conditions (i.e.\ local environmental
context provided by abiotic and potentially other biotic organisation)
and laws of physics to determine the resulting action of
the organisation in its environment.\footnote{Similarly, the view of
  the process of Generation discussed here could be elaborated into
  distinct processes of instantiation of a organism followed by a
  process of self-maintenance, where both of these are potentially
  subject to the laws of physics of the system and the local biotic
  and abiotic context.} 
Such an approach would make explicit the distinction between novelties
in Organisation Space and novelties in Action Space---the importance
of which has recently been highlighted by G.\ Wagner in his analysis
of evolutionary innovations \cite{Wagner:Evolutionary}.

Under this view, a \emph{behaviour} can be defined as the change in
state of one organisation brought about by the action of another
organisation in the environment,\footnote{We also allow the
  organisation being acted upon to be the same as that doing the
  acting, giving self-directed behaviour. Note that this is a general view of
  action and behaviour that is not confined to organisms but could be
  applied to \emph{any} organisation of matter, including subsystems
  within organisms or even abiotic organisations.} where the \emph{action} of
an organisation is the result of the application of the global laws of
physics upon it within the context of boundary conditions provided by
the local biotic and abiotic environment.\footnote{This view can be
  further expanded to allow for elaborate evolved organisations that
  control their local context to produce very reliable behaviours (we
  might call these \emph{contrivances}), and for hierarchical
  organisations comprising multiple contrivances interacting with each
  other. The distinction between organisation, local context and
  behaviour is just as relevant within a single individual as it is in
  terms of an individual's interactions with the external world.}
Biological function can then be seen as \emph{purposeful} behaviour, 
where \emph{purpose} comes about through evolutionary selection upon
evolving organisations.\footnote{This view is inspired by Pattee's
  treatment of laws, initial conditions, measurements and semantic
  closure when discussing living systems \cite{Pattee:Evolving,
    Pattee:ALife}.}

To give a concrete example in a mechanical context, imagine that we
have one organisation 
comprising a wound spring and a cogwheel both attached to an axle,
situated close to another organisation comprising a cogwheel attached
to another axle. The position of the two organisations is such that
the teeth of the two cogwheels interlock. Then, as the spring unwinds
under the laws of mechanics, it induces a rotational behaviour in the
first organisation (under the boundary condition of a fixed
translational position provided by its axle). Furthermore, this also
induces a rotational behaviour in the second organisation (which
rotates under its own axle boundary condition).

An advantage of this extended approach is that it makes explicit the
processes involved in door-opening events, as exemplified by the
red flashes in Figures~\ref{fig-typesOfOE}(b) and
\ref{fig-typesOfOE}(c). These involve
first
the discovery of a special
new state in the current Organisation Space, which then produces novel
behaviours in Action Space.
New behaviours representing expansive or transformational novelties
can arise if the
organisation is able to utilise a new feature of the laws of physics
that was not previously exploited, or if the organisation represents a new
boundary condition or contrivance that reliably generates behaviours
that would have been very unlikely before the appearance of the new
organisation.\footnote{It is also possible that a new behaviour could
  arise simply by two existing organisations being brought into a new
  relationship with each other such that they represent new boundary
  conditions for each others' actions (e.g.\ two existing organisms
  carrying out actions that they were already capable of, but doing so
  in proximity to each other such that their actions affect each other
  in a new way).}

To extend the previous mechanical example, a transformational novelty
could arise in our cogwheel system if one or both of the organisations is
augmented with a cam and follower rod that hits a metal
sheet when at its full reach; the rotation of the cam would convert the
rotational movement of the axle to a discrete linear movement in the
follower, causing the follower to hit the metal sheet at its full
reach. The result would be a discrete regular sound of the rod hitting
the metal sheet, which would represent a transformational novelty in
the system (assuming our model did not already include the concept of
discrete sound generation).

By explicitly representing behaviour as an interaction between two (or
more) organisations brought about by the action of the laws of physics
together with boundary conditions provided by the local biotic and
abiotic context, this view can accommodate Kaufmann \emph{et al}.'s notion of
\emph{unprestatability} \cite{Longo:NoEntailing}---for a given organisation or
structure, any number of behaviours are possible depending upon the
application of the laws of physics upon it in a specific local context.

The view set out in this paper and in the suggested extensions just
discussed offers a new perspective on open-ended evolution---one that
fundamentally comprises just two essential processes: the ongoing
exploration of a phenotype space (as exemplified in
Figure~\ref{fig-typesOfOE}(a) and Figure~\ref{fig-routes}), and the
discovery of door-opening states in that 
space that open up an expanded phenotype space (as exemplified by the
red flashes in Figures~\ref{fig-typesOfOE}(b) and \ref{fig-typesOfOE}(c)).
The former involves many established areas of theoretical biology
as illustrated in Figure~\ref{fig-routes} and in the accompanying
discussion, and the latter relates to the emerging topic in the
biological literature of \emph{evolutionary innovation} (e.g.\
\cite{Hochberg:Innovation}, \cite{deVladar:GrandViews}). As the topic
of innovation is currently attracting growing attention from
biologists, there is rich potential for a profitable two-way exchange of
ideas between those interested in biological innovations and those
interested in open-endedness in other kinds of evolutionary system.

\section{Acknowledgements}
\footnotesize
I am grateful to Peter Turney, Alastair Channon, Adam Stanton, James
Borg, Bradly Alicea and two anonymous reviewers for their comments and
discussions which have helped in developing the ideas set out in this
paper. 

\footnotesize
\bibliographystyle{apa-good}
\bibliography{taylor-oee-alj}

\end{document}